\documentclass[conference]{IEEEtran}
\usepackage{times}

\usepackage[numbers]{natbib}
\usepackage{multicol}
\usepackage[bookmarks=true]{hyperref}

\IEEEoverridecommandlockouts

\usepackage[utf8]{inputenc} 
\usepackage[T1]{fontenc}    
\usepackage{amsfonts}       
\usepackage{xcolor}         
\usepackage{amsmath, amssymb, bm, xargs, graphicx, booktabs}
\usepackage{url}
\usepackage{algorithmic, xfrac}
\usepackage{algorithm2e}

\definecolor{puckcolor}{RGB}{233, 113, 50}
\definecolor{malletcolor}{RGB}{0, 32, 96}

\definecolor{diagramblue}{RGB}{163, 199, 215}
\definecolor{diagramgreen}{RGB}{187, 212, 178}

\DeclareMathOperator*{\argmin}{arg\,min}

\newcommand{\trsp}{{\scriptscriptstyle\top}}

\newcommand{\rssparagraph}[1]{{\bf #1.}}

\begin{document}







\title{
Distilling Contact Planning for\\Fast Trajectory Optimization in Robot Air Hockey
}

\author{\authorblockN{Julius Jankowski\authorrefmark{1}\authorrefmark{2}\authorrefmark{3},
Ante Mari\'c\authorrefmark{1}\authorrefmark{2}\authorrefmark{3},
Puze Liu\authorrefmark{4}\authorrefmark{5}, 
Davide Tateo\authorrefmark{4},
Jan Peters\authorrefmark{4}\authorrefmark{5}\authorrefmark{6},
Sylvain Calinon\authorrefmark{2}\authorrefmark{3}}
\thanks{\authorrefmark{1}Equal contribution}
\authorblockA{\authorrefmark{2}Idiap Research Institute, Martigny, Switzerland}
\authorblockA{\authorrefmark{3}École Polytechnique Fédérale de Lausanne (EPFL), Switzerland}
\authorblockA{\authorrefmark{4}Intelligent Autonomous Systems, TU Darmstadt, Germany}
\authorblockA{\authorrefmark{5}German Research Center for AI (DFKI)}
\authorblockA{\authorrefmark{6}Centre for Cognitive Science, Hessian.AI}
\thanks{Corresponding author: \texttt{ante.maric@idiap.ch}}
}



\maketitle

\begin{abstract}
Robot control through contact is challenging as it requires reasoning over long horizons and discontinuous system dynamics.
Highly dynamic tasks such as Air Hockey additionally require agile behavior, making the corresponding optimal control problems intractable for planning in realtime. 
Learning-based approaches address this issue by shifting computationally expensive reasoning through contacts to an offline learning phase.
However, learning low-level motor policies subject to kinematic and dynamic constraints can be challenging if operating in proximity to such constraints is desired.
This paper explores the combination of distilling a stochastic optimal control policy for high-level contact planning and online model-predictive control for low-level constrained motion planning.
Our system learns to balance shooting accuracy and resulting puck speed by leveraging bank shots and the robot's kinematic structure.
We show that the proposed framework outperforms purely control-based and purely learning-based techniques in both simulated and real-world games of Robot Air Hockey.
\end{abstract}

\IEEEpeerreviewmaketitle

\section{Introduction}
\label{sec:intro}
Planning and control through non-prehensile contacts is an essential skill for robots to interact with their environment. Model-based approaches enable robots to anticipate the outcome of contact interactions given a candidate action, allowing them to find an action with the desired outcome.
Although model-based planning approaches have been shown to be successful in generating contact-rich plans for slow tasks \cite{Pang23, Jankowski2024}, highly dynamic tasks require the agent to regenerate contact plans at a sufficiently high rate to react to inherent perturbations.
These tasks have historically been used as a testbed for hardware and algorithms in robotics, with different types of games and sports, such as ball-in-a-cup~\cite{kawato1994teaching,kober2008policy}, juggling~\cite{ploeger2021high,ploeger2022controlling}, and diabolo~\cite{von2021analytical}. Dynamic tasks that involve contact, such as soccer~\cite{haarnoja2024learning}, tennis~\cite{zaidi2023athletic}, table tennis~\cite{mulling2011biomimetic,buchler2022learning}, and air hockey~\cite{liu2022robot, liu2024safe}, are typically approached with reinforcement learning methods to off-load computationally expensive reasoning through contacts to an offline exploration phase. Yet, these tasks have in common that contact with the ball or puck is instantaneous, resulting in a jump in the object state. Reasoning about the contact between the robot and the object of interest can therefore be divided into three segments of the planning horizon: \textit{i)} Moving the robot into contact, \textit{ii)} the contact itself at a single time instance, and \textit{iii)} the passive trajectory of the object after contact.

\begin{figure}[t]
    \centering
    \includegraphics[width=\linewidth]{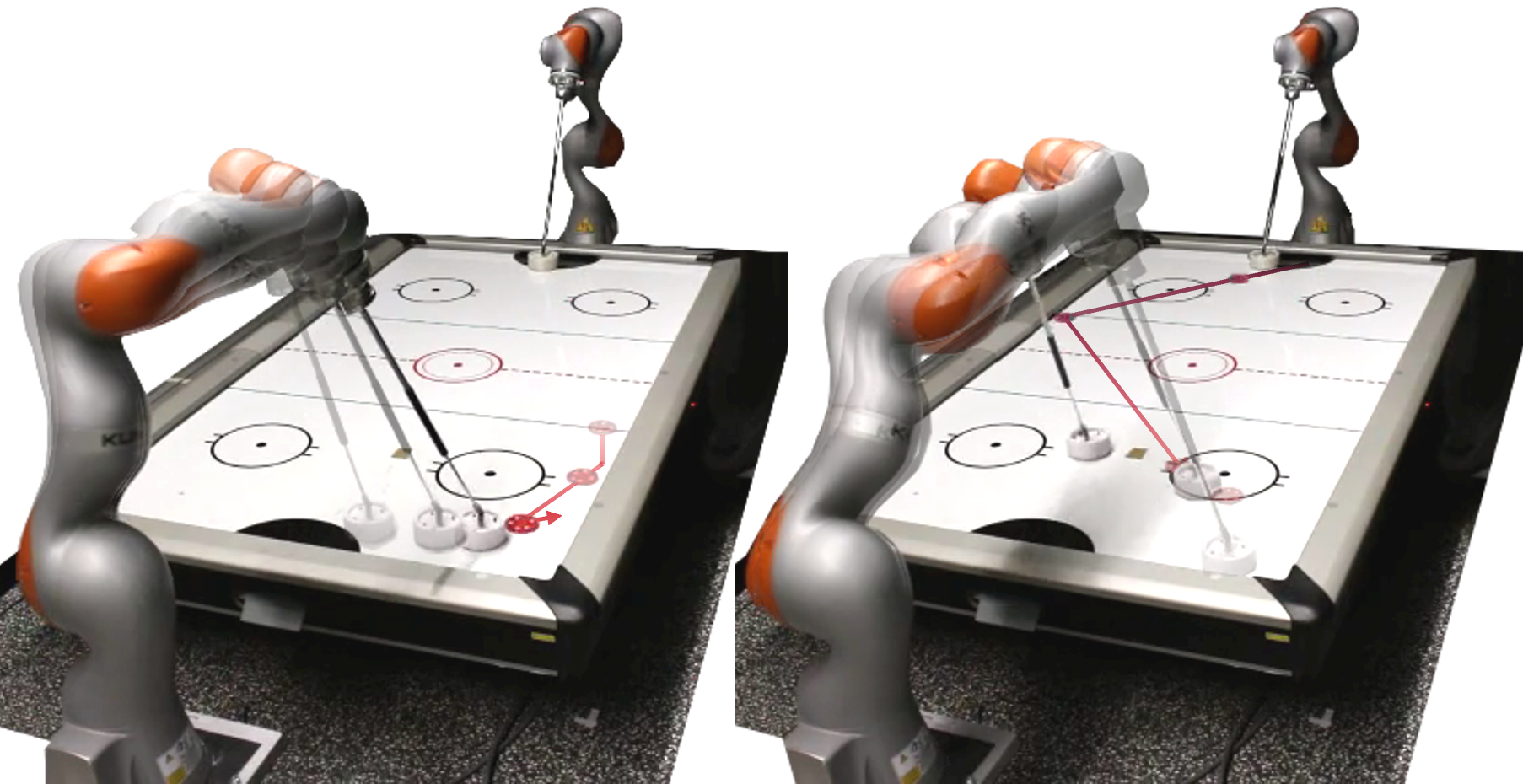}
    \caption{The proposed control framework enables our robot to autonomously play matches of air hockey. The dynamic game requires the robot to predict puck trajectories, plan the best contact, and coordinate its joints to generate high velocities without hitting a wall or lifting the mallet from the table.
    }
    \label{fig:example}
\end{figure}

This paper exploits such separability in the highly dynamic game of \textit{air hockey} (Fig. \ref{fig:example}) by combining a learning-based approach for contact planning with model-based control for moving the robot into contact. 
We show that distilling a stochastic optimal control policy enables us to effectively reduce the planning horizon required to generate desired behaviors, allowing our agent to operate in real-time. Furthermore, we highlight the interpretability of our approach by producing various shooting behaviors through different formulations of the optimal control cost, while exploiting the kinematic structure of the robot.

\begin{figure}[t]
    \centering
    \includegraphics[width=\linewidth]{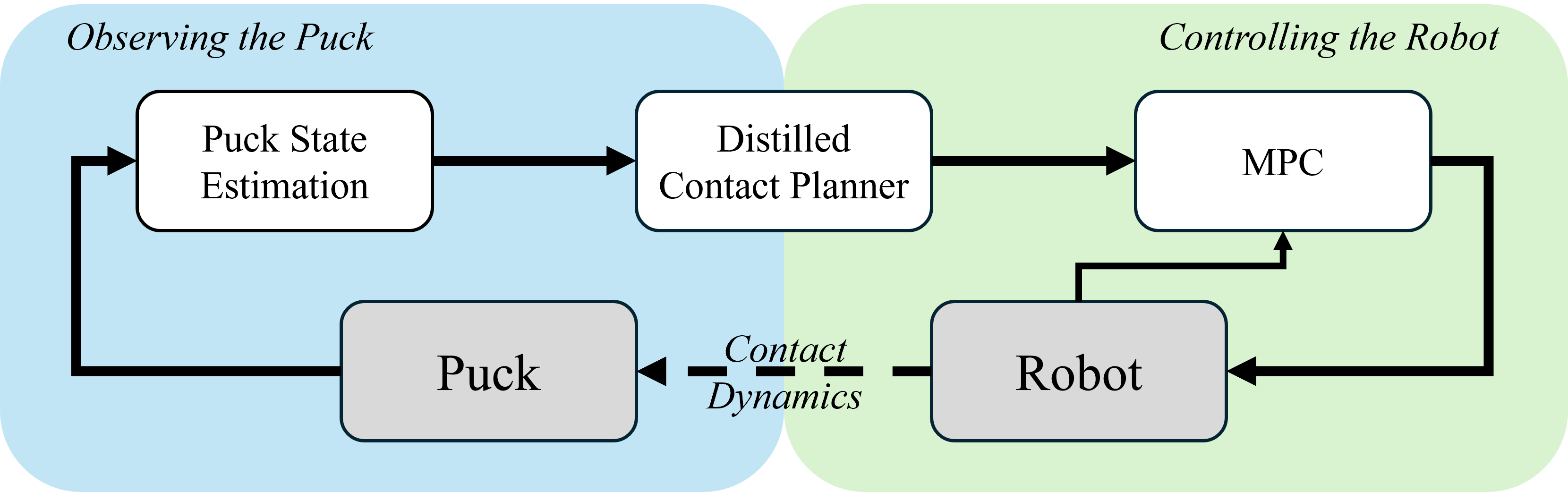}
    \caption{Overview of the interplay between puck state estimation \textcolor{diagramblue}{$\bullet$} and robot control \textcolor{diagramgreen}{$\bullet$} for closed-loop agile robot air hockey. The contact planner uses the estimated puck state to predict the puck trajectory based on the learned model. It subsequently plans a shooting angle that is used to construct an optimal control objective solved within a model-predictive controller. Robot trajectories are computed at a control rate of 50 Hz.}
    \label{fig:diagram}
\end{figure}

Fig.~\ref{fig:diagram} illustrates the online control framework that consists of state estimation, a learning-based contact planner (shooting policy), and a subsequent model-based robot controller (MPC). 
For state estimation and prediction, we learn a stochastic model of contact between the robot and the puck from data as a mixture of linear-Gaussian modes. 
Based on the learned model, we generate a dataset of example contacts that are optimal w.r.t. a stochastic optimal control objective, and distill the resulting policy through behavioral cloning. During the online phase, we retrieve optimal contact plans from the distilled policy using derivative-free inference in realtime.
Finally, for the robot controller, we use a sampling-based model-predictive controller \cite{Jankowski2023} that enforces the execution of the contact plan while respecting safety constraints such as collision avoidance with the walls.
We summarize our key contributions as follows:
\begin{itemize}
    \item We present an approach for learning the parameters of a stochastic model for discontinuous contact dynamics in robot air hockey as a mixture of linear-Gaussian modes. 
    \item We formulate contact planning for robot air hockey as a chance-constrained stochastic optimal control problem. 
    \item We propose an approach for distilling an optimal contact policy by training an implicit model to allow for planning in real-time.
\end{itemize}
After presenting our technical contributions, we provide experimental comparisons to control-based and reinforcement learning baselines. Our approach has furthermore outperformed all other approaches tested in a competitive setting~\cite{liu2024a}.

\section{Related Work}
\label{sec:sota}
The air hockey task has been part of the robotics literature for a long time~\cite{bishop1999vision}.
One of the first works using the air hockey task as a benchmark focused on skill learning of a humanoid robot~\cite{bentivegna2004learningtasks,bentivegna2004learningtoact}. In more recent years, this benchmark has been used in combination with planar robots due to high-speed motion requirements~\cite{namiki2013hierarchical,shimada2017two,igeta2017algorithm,tadokoro2022development}, and the possibility of adapting the playing style against the opponent~\cite{igeta2015decision}.
This benchmark has been recently extended to the cobot setting, where a 7-DoF robotic arm controls the mallet and maintains the table surface while striking~\mbox{\cite{alAttar2019autonomous,liu2021efficient,chuck2024robot}}.

Another use of the robot air hockey setting is as a testbed for learning algorithms. In~\cite{taitler2017learning}, deep reinforcement learning techniques are used to learn on planar robots, while in~\cite{liu2022robot}, both the planar 3-DoF and the 7-DoF cobot air hockey tasks are used to learn control policies in simulation. 
More recent techniques directly use the real 7-DoF air hockey setting as a testbed for learning algorithms: in~\cite{kicki2023fast}, the authors use learning-to-plan techniques to generate air hockey hitting trajectories in a real-world setting, while in ~\cite{liu2024safe}, this task is used to perform real-world reinforcement learning.

In general, existing solutions to the robot air hockey problem can be categorized in two main directions: learning-based approaches \cite{bentivegna2004learningtoact,taitler2017learning,liu2024safe}, and control-based approaches \cite{tadokoro2022development,alAttar2019autonomous,liu2021efficient}. Generally speaking, pure control-based approaches lead to better and faster solutions than learning-based methods but require considerable efforts in engineering and model identification, and are particularly challenging to implement and run at realtime control rates.
Instead, pure learning-based approaches obtain a lower-quality solution but make it possible to obtain more robust behaviors by relying on domain randomization and fine-tuning on the real platform. We aim to combine the advantages of learning-based and control-based approaches. We exploit both the optimality of control-based approaches for controlling the robot without considering the puck and the robustness and flexibility of learning-based approaches to efficiently generate plans for the contact between the robot and the puck to maximize the scoring probability.

\section{Learning a Stochastic Contact Model}
\label{sec:state}

Planning and controlling the contacts of the robot with the puck requires the anticipation of puck trajectories before and after contact. To enable this, we learn a simplified stochastic model of the puck dynamics for \textit{i)} estimating the current state of the puck online, \textit{ii)} predicting the trajectory of the puck online, and \textit{iii)} solving a stochastic optimal control problem to plan the next best contact between the robot and the puck.

\subsection{Mixture of linear-Gaussian Contact Dynamics}

\begin{figure}[t]
    \centering
    \includegraphics[width=\linewidth]{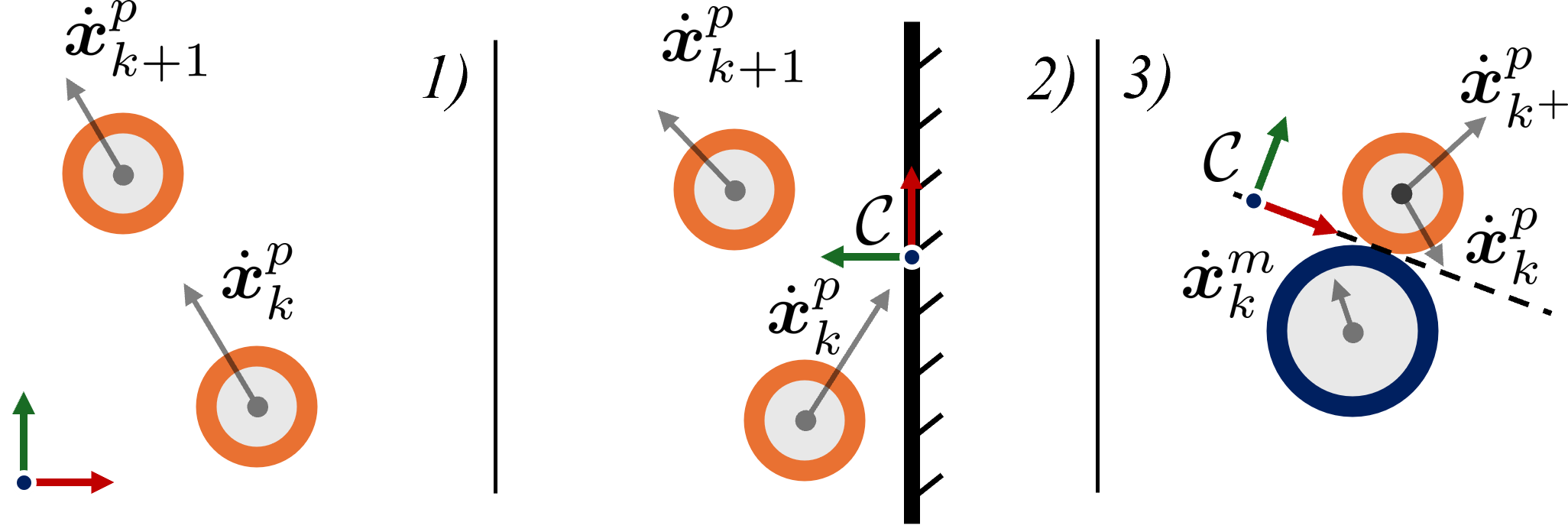}
    \caption{Illustration of three modes of the puck dynamics that are parameterized as linear-Gaussian modes. Mode \textit{1)} captures the dynamics of the puck \textcolor{puckcolor}{$\bullet$} when floating on the surface of the table. Mode \textit{2)} captures collisions between puck and walls. Mode \textit{3)} models collisions between puck and the mallet \textcolor{malletcolor}{$\bullet$} in a contact-aligned frame $\mathcal{C}$. The parameters for the nominal dynamics and the corresponding uncertainty are learned from data.}
    \label{fig:model}
\end{figure}

Suppose that $\bm{x}_k^p \in \mathbb{R}^2$ is the position of the puck w.r.t. the surface of the air hockey table at time step $k$.
The robot interacts with the puck by making contact with its mallet, i.e. the circular part of the robot's end-effector. The position of the mallet is denoted with $\bm{x}_k^m \in \mathbb{R}^2$ w.r.t. the surface of the air hockey table. We assume that the robot arm is controlled such that the mallet maintains contact with the table at all times.
In order to efficiently perform rollouts of the puck dynamics, we impose a piecewise-linear structure on the model.
Fig.~\ref{fig:model} illustrates the three modes that we present in the following: \textit{1) Floating}, \textit{2) Puck-Wall Collision}, and \textit{3) Puck-Mallet Collision}. To account for modeling errors introduced through the piecewise-linear structure, we model each mode as a conditional Gaussian distribution, resulting in a mixture of linear-Gaussian contact dynamics. In the following, we present the individual modes and their respective parameters that are learned subsequently. 

\subsubsection{Floating}
The first mode captures the dynamics of the puck when it is freely floating on the table and is not in collision with the wall or mallet.
The prediction of the puck velocity is modeled stochastically with
\begin{equation}
    \mathrm{Pr}_1\left(\dot{\bm{x}}_{k+1}^p | \dot{\bm{x}}_{k}^p\right) = \mathcal{N}\left(\bm{\Theta}_1 \dot{\bm{x}}_{k}^p + \bm{\theta}_1, \bm{\Sigma}_1\right),
\end{equation}
where $\bm{\Theta}_1, \bm{\theta}_1, \bm{\Sigma}_1$ are parameters of the conditional Gaussian distribution.

\subsubsection{Puck-Wall Collision}
The second mode models the dynamics of the puck reflecting against the wall.
The prediction of the velocity is modeled in a coordinate system $\mathcal{C}$ that is aligned with the contact surface of the corresponding wall. The one-step prediction of the puck velocity is modeled with
\begin{equation}
    \mathrm{Pr}_2\left({}^c\dot{\bm{x}}_{k+1}^p | {}^c\dot{\bm{x}}_{k}^p\right) = \mathcal{N}\left(\bm{\Theta}_2 {}^c\dot{\bm{x}}_{k}^p + \bm{\theta}_2, \bm{\Sigma}_2\right),
\end{equation}
where ${}^c\dot{\bm{x}}^p$ is the puck velocity in the contact-aligned coordinate system. $\bm{\Theta}_2, \bm{\theta}_2, \bm{\Sigma}_2$ are parameters of this mode.

\subsubsection{Puck-Mallet Collision}
As a third mode, we model the interaction between the puck and the mallet as a collision in which the velocity of the puck changes instantaneously at the time of contact. We also model this mode using a conditional Gaussian distribution
\begin{equation}
\label{eq:mode_3}
    \mathrm{Pr}_3\left({}^c\dot{\bm{x}}_{k^+}^p | {}^c\dot{\bm{x}}_{k^-}^p, {}^c\dot{\bm{x}}_{k}^m\right) = \mathcal{N}\left(\bm{\Theta}_3^p {}^c\dot{\bm{x}}_{k^-}^p \!+\! \bm{\Theta}_3^m {}^c\dot{\bm{x}}_{k}^m \!+\! \bm{\theta}_3, \bm{\Sigma}_3\right).
\end{equation}
The velocities of the puck ${}^c\dot{\bm{x}}^p$ and of the mallet ${}^c\dot{\bm{x}}^m$, respectively, are expressed in the contact-aligned coordinate system $\mathcal{C}$. The index $k^+$ corresponds to time step $k$ after applying the collision model, while $k^-$ describes the instant right before the collision. The model parameters for the third mode are $\bm{\Theta}_3^p, \bm{\Theta}_3^m, \bm{\theta}_3, \bm{\Sigma}_3$.

\subsection{Learning Model Parameters from Data}

Given recorded trajectories of the puck and the mallet, the data is fragmented into consecutive puck velocity pairs together with the mallet velocity, i.e. $\dot{\bm{x}}_{k}^p, \dot{\bm{x}}_{k+1}^p, \dot{\bm{x}}_{k}^m$, and the corresponding mode is assigned to each data sample. As a result, we assume to obtain a dataset $\{ \bm{y}_{i,n}, \bm{\xi}_{i,n} \}_{n=0}^{N_i}$ for each mode $i$, where $\bm{y}_{i,n}$ is the $n$-th velocity prediction sample for mode $i$, e.g. $\bm{y}_{1} = \dot{\bm{x}}_{k+1}^p$, and $\bm{\xi}_{i,n}$ is the $n$-th prediction condition sample for mode $i$, e.g. $\bm{\xi}_{1} = \dot{\bm{x}}_{k}^p$. To learn the parameters of the model, we fit a Gaussian distribution to the dataset for each mode modeling the joint probability distribution of prediction and condition with
\begin{equation}
    \mathrm{Pr}_i(\bm{y}_{i}, \bm{\xi}_{i}) = \mathcal{N}\left( \begin{pmatrix} \bm{\mu}_{\bm{y}_i} \\ \bm{\mu}_{\bm{\xi}_i} \end{pmatrix}, \begin{pmatrix} \bm{\Sigma}_{\bm{y}_i} & \hspace{-1em} \bm{\Sigma}_{\bm{y}_i \bm{\xi}_i} \\ \bm{\Sigma}_{\bm{y}_i \bm{\xi}_i}^\trsp & \hspace{-1em} \bm{\Sigma}_{\bm{\xi}_i} \end{pmatrix} \right).
\end{equation}
Given the parameters of the joint probability distribution, the parameters of the linear-Gaussian models can be computed by conditioning the probability distribution on the input $\bm{\xi}$. Thus, the parameters are given by
\begin{align}
    \begin{split}
        \bm{\Theta}_i &= \bm{\Sigma}_{\bm{y}_i \bm{\xi}_i} \bm{\Sigma}_{\bm{\xi}_i}^{-1},\\
        \bm{\theta}_i &= \bm{\mu}_{\bm{y}_i} - \bm{\Sigma}_{\bm{y}_i \bm{\xi}_i} \bm{\Sigma}_{\bm{\xi}_i}^{-1} \bm{\mu}_{\bm{\xi}_i},\\
        \bm{\Sigma}_i &= \bm{\Sigma}_{\bm{y}_i} - \bm{\Sigma}_{\bm{y}_i \bm{\xi}_i} \bm{\Sigma}_{\bm{\xi}_i}^{-1} \bm{\Sigma}_{\bm{y}_i \bm{\xi}_i}^\trsp.
    \end{split}
\end{align}

\subsection{Piecewise-linear Kalman Filtering}

The learned linear-Gaussian models allow us to update the estimated state of the puck using the Kalman filter. As a result, an estimate of the puck state at time step $k$, i.e. $\hat{\bm{s}}_{k} = \left( \hat{\bm{x}}_{k}^{p^\trsp}, \dot{\hat{\bm{x}}}_{k}^{p^\trsp} \right)^\trsp$, is obtained based on a noisy measurement of the puck position $\tilde{\bm{x}}_{k}^p$. For this, the mode of the dynamics is detected at each time step such that the corresponding parameters are used within the Kalman filter update. The parameters are translated into linear-Gaussian state-space dynamics, i.e.
\begin{equation}
    \mathrm{Pr}_i\left(\bm{s}_{k+1} | \bm{s}_{k}\right) = \mathcal{N}\left(\bm{A}_i \bm{s}_{k} + \bm{b}_i, \bm{Q}_i\right),
\end{equation}
with system parameters $\bm{A}_i, \bm{b}_i$ and process noise covariance matrix $\bm{Q}_i$ computed with
\begin{align}
    \begin{split}
        \bm{A}_i = \begin{pmatrix} \bm{A}_{i, \bm{x}\bm{x}} & \bm{A}_{i, \bm{x}\dot{\bm{x}}} \\ 
                                   \bm{0} & \bm{\Theta}_i\end{pmatrix}; \;
        \bm{b}_i = \begin{pmatrix} \bm{0} \\ \bm{\theta}_i\end{pmatrix}; \;
        \bm{Q}_i = \begin{pmatrix} \bm{0} & \bm{0} \\ 
                                    \bm{0} & \bm{\Sigma}_i\end{pmatrix}.
    \end{split}
\end{align}
Here, the model parameters $\bm{A}_{i, \bm{x}\bm{x}}, \bm{A}_{i, \bm{x}\dot{\bm{x}}}$ determine the prediction of the puck position at the next time step given the current puck position and velocity. These parameters are derived using numerical integration and are constant.

\subsection{Probability of Hitting the Goal}
\label{sec:prob_goal}

For the robot to anticipate whether a candidate shot may lead to scoring a goal, we predict the probability of hitting the goal based on the learned linear-Gaussian puck dynamics. Note that the probability of hitting the goal does not account for a defending opponent. Without loss of generality, suppose that the collision between mallet and puck happens at $k=0$. Given the puck state at the time of collision $\hat{\bm{s}}_{0^-}$ and the corresponding mallet state $\bm{x}_{0}^m, \dot{\bm{x}}_{0}^m$, the expected puck velocity after the collision is computed as defined in \eqref{eq:mode_3}, resulting in the expected puck state $\hat{\bm{s}}_{0^+}$. By rolling out the discretized stochastic model with
\begin{align}
\label{eq:stochastic_rollout}
    \begin{split}
        \hat{\bm{s}}_{k+1} &= \bm{A}_{i_k} \hat{\bm{s}}_{k} + \bm{b}_{i_k},\\
        \bm{P}_{k+1} &= \bm{A}_{i_k} \bm{P}_{k} \bm{A}_{i_k}^\trsp + \bm{Q}_{i_k},
    \end{split}
\end{align}
a Gaussian distribution of puck states, i.e. $\bm{s}_k \sim \mathcal{N}(\hat{\bm{s}}_k, \bm{P}_k)$ is obtained for each time step $k>0$. The rollout is initialized with $\hat{\bm{s}}_{0} = \hat{\bm{s}}_{0^+}$ and $\bm{P}_{0} = \bm{Q}_{3}$, exploiting the separated stochastic model of collisions between mallet and puck.

\label{sec:goal_prob}
\begin{figure}[t]
    \centering
    \includegraphics[width=\linewidth]{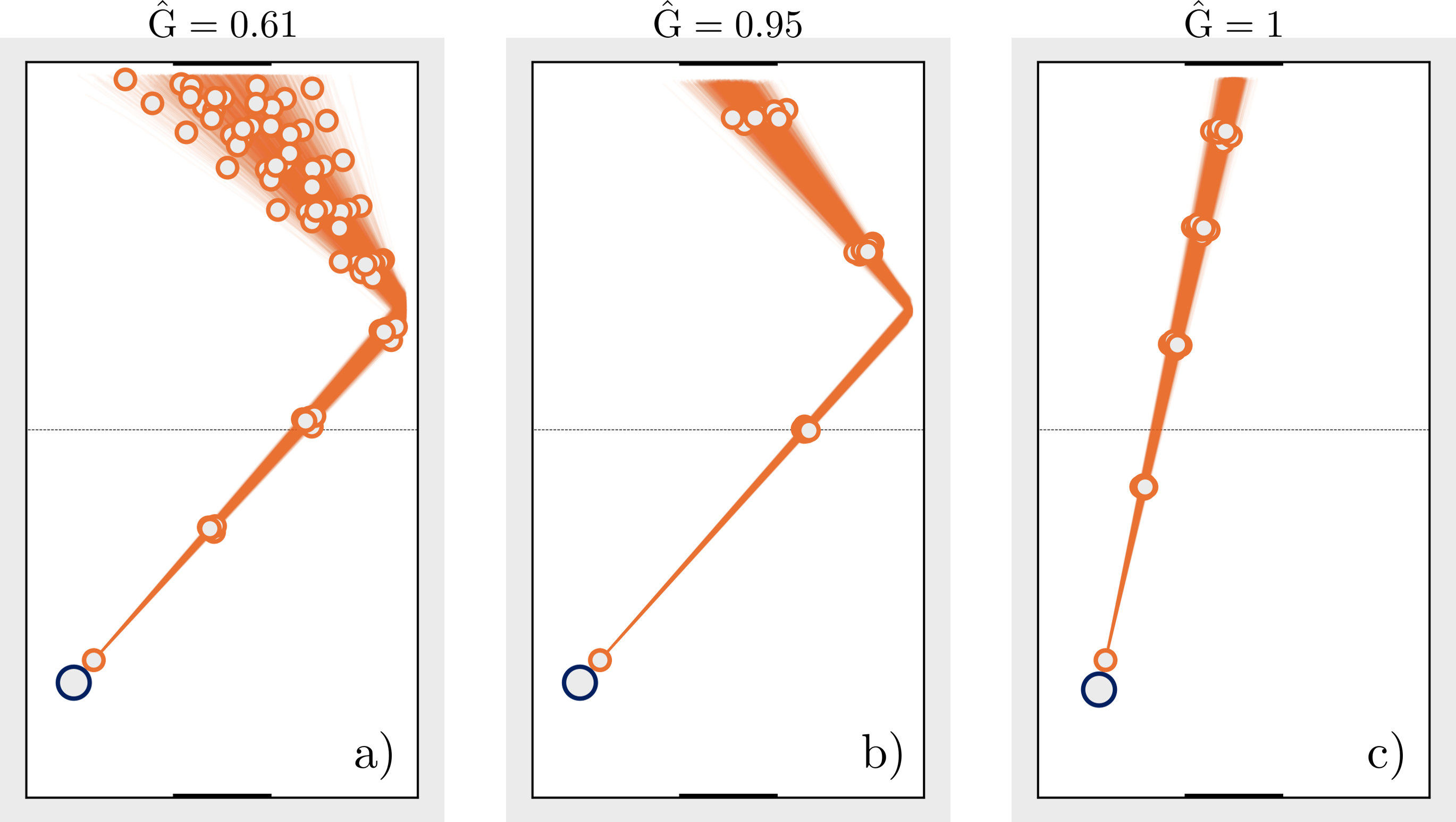}
    \caption{A qualitative comparison of the probability of hitting the goal $\hat{\mathrm{G}}$ for different shooting angles and shooting speeds. The shooting angles are indicated by the mallet position \textcolor{malletcolor}{$\bullet$} w.r.t. the puck position \textcolor{puckcolor}{$\bullet$} at the time of contact. The shooting speed, i.e. the speed of the mallet at the time of contact, is 1.2~$\frac{\mathrm{m}}{\mathrm{s}}$ for a) and c), while the shooting speed is 2~$\frac{\mathrm{m}}{\mathrm{s}}$ for b).}
    \label{fig:rollout}
\end{figure}

To evaluate the probability of scoring a goal, we perform the stochastic rollout as defined in \eqref{eq:stochastic_rollout} until the expected puck position $\hat{\bm{x}}^p_k$ crosses the goal line. We denote this time step with $k_{\mathrm{goal}}$. In the following, we denote the probability of scoring a goal, i.e. $\mathrm{G}=1$, given a puck position as a Bernoulli distribution with
\begin{equation}
    \mathrm{Pr}\left(\mathrm{G}=1 | \bm{x}^p_{k}\right) = \begin{cases} 1, \quad \mathrm{if}\; \bm{x}^p_{k} \in \mathcal{X}_{\mathrm{goal}} \\ 0, \quad \mathrm{else.} \end{cases}
\end{equation}
The subset in puck position space $\mathcal{X}_{\mathrm{goal}}$ represents the goal region. Consequently, we can compute the probability of scoring a goal given the initial conditions of a shot by marginalizing over the puck position at time step $k_{\mathrm{goal}}$ with
\begin{equation}
\label{eq:prob_scoring}
    \mathrm{Pr}\left(\mathrm{G}=1 | \hat{\bm{s}}_{0^-}, \bm{x}_{0}^m, \dot{\bm{x}}_{0}^m\right) = \int_{\mathcal{X}_{\mathrm{goal}}} \mathrm{Pr}\left( \bm{x}^p_{k_{\mathrm{goal}}} \right) d \bm{x}^p_{k_{\mathrm{goal}}}.
\end{equation}
We compute the probability in \eqref{eq:prob_scoring} using Monte-Carlo approximation by sampling $N_\mathrm{G}$ puck positions from the Gaussian distribution at prediction time step $k_{\mathrm{goal}}$ and counting the number of samples that would hit the goal
\begin{equation}
\label{eq:prob_scoring_mc}
    \mathrm{Pr}\left(\mathrm{G}=1 | \hat{\bm{s}}_{0^-}, \bm{x}_{0}^m, \dot{\bm{x}}_{0}^m\right) \approx 
\frac{1}{N_\mathrm{G}} \sum_{n=1}^{N_\mathrm{G}} \mathrm{Pr}\left(\mathrm{G}=1 | \bm{x}^p_{k_{\mathrm{goal}}, n}\right),
\end{equation}
with $\bm{x}^p_{k_{\mathrm{goal}}, n} \sim \mathrm{Pr}( \bm{x}^p_{k_{\mathrm{goal}}} )$. In the following, we denote the approximated probability of hitting the goal, corresponding to the right-hand side of \eqref{eq:prob_scoring_mc}, with $\hat{\mathrm{G}}$.

Fig.~\ref{fig:rollout} illustrates stochastic rollouts for various initial conditions of a shot. Evaluating $\hat{\mathrm{G}}$ as defined in \eqref{eq:prob_scoring_mc}, we observe that those initial conditions have a significant effect even if the expected puck trajectory hits the center of the goal for all conditions. Fast shots (Fig.~\ref{fig:rollout}-b) accumulate less uncertainty compared to slow shots (Fig.~\ref{fig:rollout}-a) since the modeled process noise is constant over time. Direct shots accumulate less uncertainty during rollout than bank shots, as collisions with a wall add significant process noise (Fig.~\ref{fig:rollout}-c).

\section{Implicit Contact Planning under Uncertainty}
\label{sec:planner_control}
\begin{figure*}
    \centering
    \includegraphics[width=\linewidth]{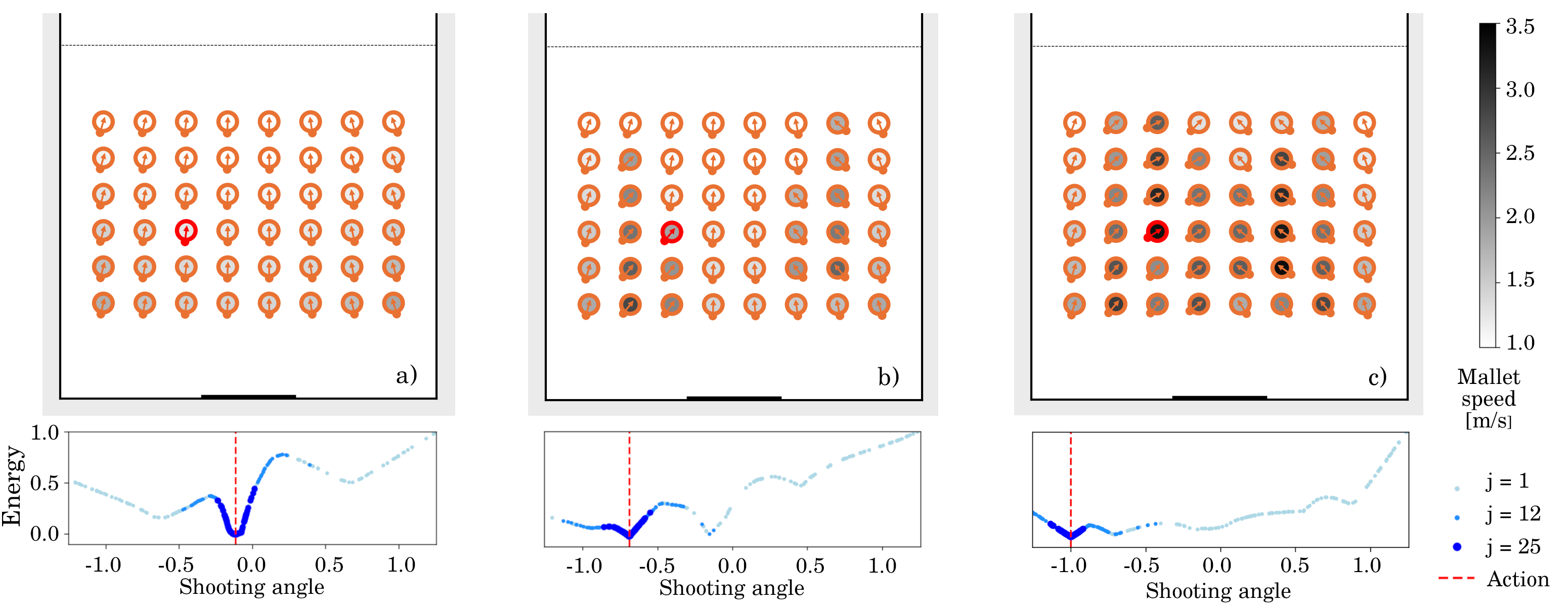}
    \caption{Examples of differently tuned shooting plans and corresponding energy landscapes. Shooting direction and mallet speed are displayed for varying initial puck positions. Instance a) evaluates only scoring probability ($\lambda_1=1,~\lambda_2=0,~\beta=0.5$); b) adds additional weight on expected puck speed at the goal line ($\lambda_1=1,~\lambda_2=0.2,~\beta=0.5$); c) evaluates only the expected puck speed at the goal line ($\lambda_1=0,~\lambda_2=1,~\beta=0.5$). The energy landscape and sampling process at timesteps $j\in\{1,12,25\}$ are visualized for an example shot denoted in red. All pucks are static at $j=0$.}
    \label{fig:sampling}
\end{figure*}

The learned dynamics model enables the prediction of uncertain puck trajectories for contact planning. In particular, we search for contact states of the mallet that result in desired puck trajectories after contact. The proposed contact planning module is based on stochastic optimal control, optimizing the mallet's state at contact. We combine this optimization with a model-based robot controller driving the robot to the desired contact state at the desired time (cf. Fig. \ref{fig:diagram}). 

\subsection{Stochastic Optimal Control for Shooting}
\label{subsec:soc}

Given the desired time of contact and the corresponding estimate of the puck state $\bm{s}_{0^-}$ at that time, we pose contact planning for shooting as a stochastic optimal control problem searching for the mallet state $\bm{x}_{0}^m, \dot{\bm{x}}_{0}^m$ at the time of contact. For this, we aim to maximize a tradeoff between the probability of hitting the goal $\hat{\mathrm{G}}$ and the expected puck speed $v_{\mathrm{puck}}$ at the goal line. The expected puck speed is computed as the norm of the mean puck velocity at $k_{\mathrm{goal}}$ according to Sec.~\ref{sec:goal_prob}. While the probability of hitting the goal $\hat{\mathrm{G}}$ does not account for a defending opponent, we use the speed of the puck as a measure of the difficulty of defending against the shot. The stochastic optimal control problem is given as
\begin{align}
\begin{split}
\label{eq:soc}
    \max_{\bm{x}_{0}^m, \dot{\bm{x}}_{0}^m} \quad &\lambda_1 \hat{\mathrm{G}} + \lambda_2 v_{\mathrm{puck}}\\
    \mathrm{s.t.} \quad &\hat{\mathrm{G}} > \beta,
\end{split}
\end{align}
where we deploy an additional chance constraint to enforce the probability of hitting the goal to be higher than a threshold $\beta$ based on the learned stochastic model. The weights $\lambda_1$ and $\lambda_2$ are used for tuning for the desired behavior. Based on the qualitative comparison illustrated in Fig.~\ref{fig:rollout}, we expect that solely optimizing for the probability of hitting the goal results only in direct shots, as bank shots induce uncertainty. Yet, due to the kinematics of the robot, the puck speed may be increased with bank shots. Thus, depending on the puck state, the tuned objective can produce both straight shots and bank shots, increasing the chances of scoring.

\subsection{Shooting Angle as Reduced Action Space}
\label{subsec:action_parametrzation}

The goal of the shooting policy is to find the optimal mallet state at the time of contact, i.e. $\bm{x}_{0}^m$ and $\dot{\bm{x}}_{0}^m$, respectively. Due to the underlying contact geometry and constraints, for a given puck position $\hat{\bm{x}}^p_0$ we parameterize the mallet position as a shooting angle $u \in \mathcal{U}$ between the mallet and puck. Note that a shooting angle of $u = 0$ corresponds to a straight shot that is parallel to the side walls of the table. We reduce the dimensionality of the action space further by imposing two heuristic constraints on the mallet velocity $\dot{\bm{x}}_{0}^m$: \textit{i)} The mallet velocity at the time of contact aligns with the shooting angle, such that $\dot{\bm{x}}_{0}^m = v (\cos u, \sin u)^\trsp$
with scalar velocity $v > 0$ encoding the norm of the mallet velocity. While this constraint excludes shooting angles that are not aligned with the mallet velocity, it enforces maximum transmission of kinetic energy from the robot to the puck. \textit{ii)} We impose that the norm of the mallet velocity is maximal given a shooting configuration $\bm{q}_0$ of the robot and velocity limits $\mathcal{\dot{Q}}$ of the joints of the robot, such that
\begin{align}
\begin{split}
    v^* = &\max v\\
    \text{s.t.} \quad &v \bm{e}_u = \bm{J}(\bm{q}_0) \dot{\bm{q}}_0,\\
    &\dot{\bm{q}}_0 \in \mathcal{\dot{Q}}.
\end{split}
\end{align}
Note that the unit vector $\bm{e}_u \in \mathbb{R}^3$ encodes the shooting direction including zero contribution in the z-direction. Accordingly, $\bm{J}(\bm{q}_0) \in \mathbb{R}^{3 \times n_{\text{dof}}}$ corresponds to the Jacobian w.r.t. the Cartesian position of the mallet.

As a result, the shooting angle $u$ is the action that we optimize for. With the imposed constraints, a shooting angle uniquely maps to a mallet state at the time of contact. Thus, in the following, we denote the probability of scoring a goal as a function of the shooting angle with $\mathrm{Pr}\left(\mathrm{G}=1 | \hat{\bm{s}}_{0^-}, u\right)$.

\subsection{Distilling an Optimal Contact Policy}
\label{subsec:EBM_train}

The long-horizon predictions required to plan shooting actions make it difficult to operate at a rate that is sufficient for agile behavior. We address this challenge by training an implicit model to generate solutions to the stochastic optimal shooting problem \eqref{eq:soc} in realtime. Since the stochastic policy we aim to capture is multimodal, we opt for an implicit policy representation due to its ability to learn multimodal action distributions \cite{Florence2022, Chi2023}. In the case of air hockey, example modes correspond to the number of reflections against the bank of the table, ranging from straight shots without reflections to bank shots with multiple reflections. To learn the implicit policy, we use an energy-based model (EBM) \cite{Florence2022}. Access to the learned energy\footnote{In this paper, the term \textit{energy} refers to the negative logarithm of the unnormalized probability density function that we want to model.} landscape allows us to accommodate various sampling strategies without retraining, enabling a balance between optimality and computational efficiency.

We train the EBM by solving the computationally expensive shooting angle optimization offline and using the results as training data. Namely, we first generate a dataset of shooting angles for $N$ different puck states at time of contact $\{\bm{s}_{i}\}_{i=1}^{N}$.
Due to the one-dimensional parametrization of the action space introduced in Sec.~\ref{subsec:action_parametrzation}, we efficiently explore the space of shooting angles for each initial puck state by sampling $M$ candidate shooting angles $\{u_{i}^{j}\}_{j=1}^{M}$. Subsequently, we compute the stochastic rollout of the puck trajectory as presented in Sec.~\ref{sec:prob_goal} and evaluate the objective and chance constraint from \eqref{eq:soc}. 
The best-performing sample $\hat{u}_i$ is then used as a positive example for training the implicit behavioral cloning model, with the remaining $M-1$ samples as negative counter-examples. This results in a dataset of $M \times N$ state-action pairs $\{\bm{s}_{i}, \hat{u}_i, \{u_{i}^{j}\}_{j=1}^{M-1}\}_{i=1}^{N}$, 
which we use to train the EBM $E_\theta(\bm{s}, u)$ using an InfoNCE-style \cite{Oord2018RepresentationLW} loss 
\begin{align}
\label{eq:ebm_training1}
\mathcal{L}_\text{InfoNCE}&=\sum_{i=1}^N-\log\left(\Tilde{p}_\theta\big(\hat{u}_i|\bm {s}_{i}, \{u^j_i\}_{j=1}^{M-1}\big)\right).
\end{align}
In the above, counter-examples $\{u^j_i\}_{j=1}^{M-1}$ are used to compute the likelihood $\Tilde{p}_\theta$ as follows:
\begin{align}
\label{eq:ebm_training2}
\Tilde{p}_\theta\big(\hat{u}_i|\bm {s}_{i}, \{u_j^i\}_{j=1}^{M{-}1}\big) &= \frac{e^{-E_\theta(\bm {s}_{i}, \hat{u}_i)}}{e^{-E_\theta(\bm {s}_{i}, \hat{u}_i)} + \sum_{j=1}^{M-1}e^{-E_\theta(\bm {s}_{i}, u_j^i)}}.
\end{align}
The described loss function reduces energy $E_\theta(\bm{s}, u)$ for shooting angles that solve the optimization problem in \eqref{eq:soc}, while increasing the energy of non-optimal shooting angles. Once the model is trained, this allows us to infer optimal shooting angles using sampling-based optimization.

\subsection{Online Inference with Warm-Starting}
\label{subsec:EBM_infer}

\RestyleAlgo{ruled}
\SetKwComment{Comment}{/* }{ */}
\SetKw{IN}{Input:}

\SetKw{IN}{Input:}
\begin{algorithm}[t!]
\DontPrintSemicolon
\caption{Shooting policy (EBM inference)}
\label{alg:sampling_actions}
\hspace{-.9em}\KwIn{Puck state $\hat{\bm{s}}_{0^-}$, variance $\sigma$, samples $\left\{\Tilde{p}_{i}, \Tilde{u}_{i}\right\}_{i=1}^{N}$}
\hspace{-.9em}\KwOut{Shooting angle $\hat{u}$, new samples $\left\{\Tilde{p}_{i}, \Tilde{u}_{i}\right\}_{i=1}^{N}$}\vspace{.1em}
$\{\tilde{u}_{i}\}_{i=1}^{N} \gets \sim \text{Multinomial}(N, \{\tilde{p}_{i}\}_{i=1}^{N}, \{\tilde{u}_{i}\}_{i=1}^{N})$\;
$\{\tilde{u}_{i}\}_{i=1}^{N} \gets \{\tilde{u}_{i}\}_{i=1}^{N} + \sim \mathcal{N}(0, \sigma)$\;
$\{\tilde{u}_{i}\}_{i=1}^{N} \gets$ clip $\{\tilde{u}_{i}\}_{i=1}^{N}$ to $\mathcal{U}$\;
$\{E_{i}\}_{i=1}^{N} \gets \{E_\theta(\hat{\bm{s}}_{0^-}, \tilde{u}_{i})\}_{i=1}^{N}$\;
$\{\tilde{p}_{i}\}_{i=1}^{N} \gets \text{softmax}(-\{E_{i}\}_{i=1}^{N})$\;
$\hat{u} \gets \text{argmax}(\{\tilde{p}_{i}\}, \{\tilde{u}_{i}\})$\;
\end{algorithm}

To solve \eqref{eq:soc} given the estimated puck state $\hat{\bm{s}}_{0^-}$, we search for a state-action pair that minimizes the learned energy, i.e.
\begin{align}
\label{eq:ebm_inference}
\hat{u}=\argmin_{u\in \mathcal{U}}{E_\theta(\hat{\bm{s}}_{0^-}, u)}.
\end{align}
For realtime optimization, we leverage direct access to the learned energy landscape of the EBM by executing sampling iterations concurrently with other components of the control loop. This allows us to refine contact plans as the robot executes trajectories, while considering only the shrinking horizon from the current timestep to the time of contact.
We base the online retrieval of optimal shooting angles on derivative-free optimization procedures from \cite{Florence2022}. As in the offline scenario, shooting actions are generated online by uniformly sampling $N$ candidate actions, inferring their energy values, and resampling with replacement to warm-start optimization in each subsequent timestep.
To converge towards a solution with minimum implicit energy, reductions to the sampling variance 
are applied at each timestep, while keeping the optimal contact angle $\hat u$ as a reference for the mid-level trajectory planner. A full iteration of EBM inference is outlined in Alg. \ref{alg:sampling_actions}. 
We observe that the learned energy models and utilized optimization procedure efficiently retrieve multimodal contact plans to produce desired behaviors, as shown in Fig. \ref{fig:sampling}.

\section{Experimental Evaluation}
\label{sec:exp}

This section details the simulated and real-world experiments used to validate our approach in an online contact planning setting. We evaluate the shooting performance of a robot arm controlled by our framework and compare it against state-of-the-art 
approaches for robot air hockey.


\subsection{Implementation Details}

\subsubsection{Data Collection for Puck Dynamics} 
We use data collected in a physics-based simulator to learn model parameters of puck dynamics as presented in Sec.~\ref{sec:state}. One set of data is collected by randomly moving the robot's end-effector into contact with the puck and the other set of data is collected without moving the robot and by initializing the puck with a high random velocity. In total, the training set consists of 100 episodes with 50 time steps each, which corresponds to a total of 100 seconds of observations of the puck dynamics.

\subsubsection{EBM Architecture and Training}
\label{sec:EBM_details}
The energy-based shooting model consists of a multilayer perceptron with $2$ hidden, fully connected layers of $128$ neurons each. The model is trained on $N\!=\!3000$ initial puck states, with $M\!=\!100$ action samples. For each action sample, the corresponding state action pair is evaluated using the stochastic dynamics model learned from simulated data (Sec.~\ref{sec:state}), and the optimal control cost (Sec.~\ref{sec:planner_control}).
Training required $500$ epochs to converge for satisfactory performance using the Adam \cite{kingma2014adam} optimizer with a decaying learning rate.

\subsection{Experimental Setup}


\begin{figure}
    \centering
    \includegraphics[width=1.\linewidth]{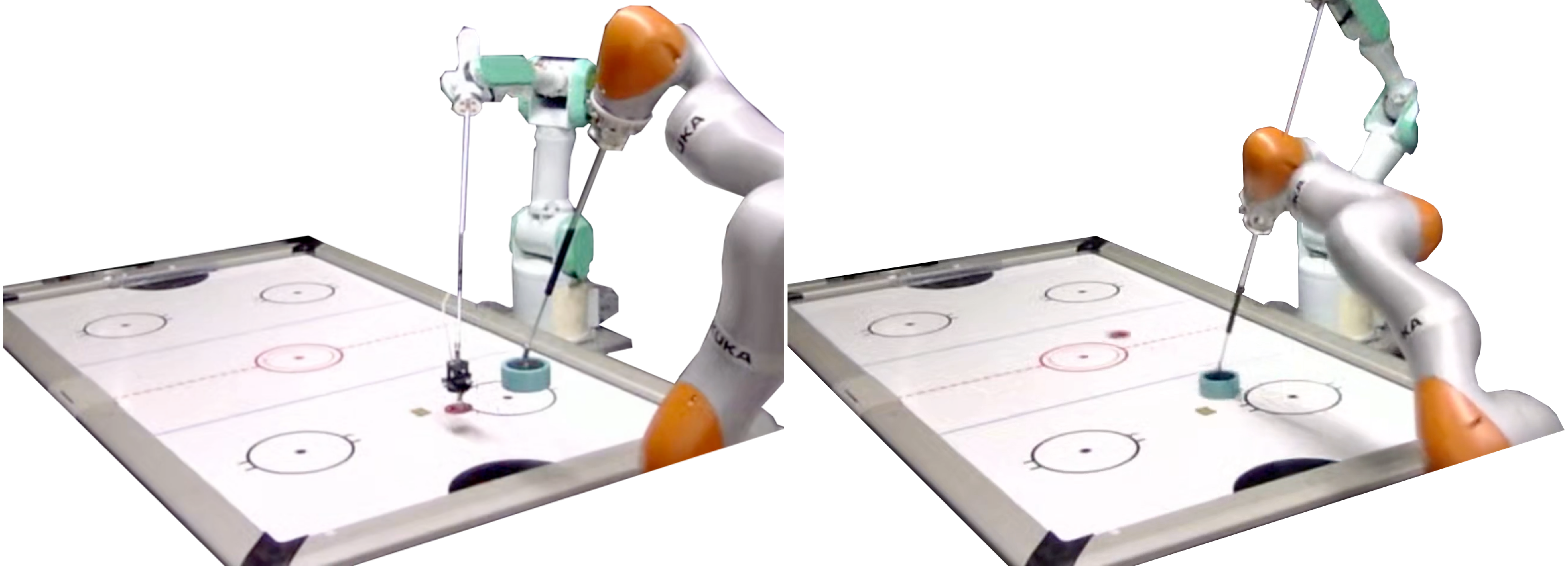}
    \caption{The automated experimental setup consists of a robot placing the puck at a pre-defined grid of positions (left image). After releasing the puck, the KUKA robot executes the shooting policy (right image). We measure the speed of the puck at the goal line and whether the puck hit the goal as quality metrics.}
    \label{fig:exp_setup}
\end{figure}

The experiment is conducted with a KUKA iiwa14 LBR manipulator equipped with a mallet end-effector that is attached to a passive joint for seamless contact with the table surface. Experiments are carried out in a simulated \textit{MuJoCo} environment and on a real-world setup (cf. Fig. \ref{fig:exp_setup}) for evaluation of \textit{sim2real} transfer. We evaluate three instances of the proposed approach by using different parameters for the chance-constrained optimization problem in \eqref{eq:soc}. \textbf{Ours \#1}: a \textit{conservative} policy that prioritizes accuracy ($\lambda_1\!=\!1, \lambda_2\!=\!0, \beta\!=\!0.5$); \textbf{Ours \#2}: a \textit{balanced} policy that compromises between accuracy and puck speed ($\lambda_1\!=\!1, \lambda_2\!=\!0.2, \beta\!=\!0.5$); and \textbf{Ours \#3}: an \textit{aggressive} policy that prioritizes puck speed ($\lambda_1\!=\!0, \lambda_2\!=\!1, \beta\!=\!0.5$). Simulated results are also illustrated in Fig.~\ref{fig:sampling} for initial puck velocities of zero. We compare the three instances of our contact planner with: \textbf{CB}, a baseline that utilizes conventional planning and control methods \cite{liu2021efficient}; and \textbf{ATACOM}, a safe reinforcement learning approach for learning a robot policy \cite{liu2022robot}. Note that the ATACOM policy is trained in simulation and deployed in the physical experiment without additional tuning or retraining.

We perform 100 shots with each policy and report the accuracy score, puck speed at the goal line, and the number of bank reflections for successful shots. 
Each shot is initialized by placing the puck within a grid in front of the robot. Due to imperfect air flow on the air hockey table, the puck moves after release, requiring the robot to adapt for a good shot.

\subsection{Results}

Recorded metrics are reported in Table~\ref{tab:sim_res} for the simulated environment and in Table~\ref{tab:phys_res} for the real-world environment. Compared to \textbf{CB} and \textbf{ATACOM}, we observe that our framework is capable of achieving higher scoring accuracy and significantly higher puck speeds in both environments. It can be seen that different instances of our policy obtain either a high score or high puck speeds according to the corresponding parameters of the stochastic optimal control problem. For example, when compared to \textbf{Ours \#1}, it can be seen that \textbf{Ours \#3} compromises scoring accuracy for faster puck speeds and a high number of bank reflections, potentially making the shots more difficult to defend against. The higher number of bank reflections produced by \textbf{Ours \#3} indicates that the robot kinematics allow for higher shooting speeds when hitting laterally, at the risk of missing the goal due to uncertainty gained with every bank reflection. Note that the score of \textbf{Ours \#3} is lower than the score chance threshold $\beta=0.5$ for this instance. This indicates that the learned model either has an error in the nominal dynamics or expects too little uncertainty gain due to bank reflections. We further note a decrease in performance for all agents due to the \textit{sim2real} gap, with \textbf{ATACOM} showing the highest sensitivity to transfer as it requires fine-tuning on the real environment. \textbf{CB} shows the least decrease in performance, as it is parameterized for the real system. However, note that the shooting trajectory optimization loop of \textbf{CB} is slower than required to run at 50 Hz, making it prone to errors due to the puck moving unpredictably during the shooting motion. 
Since we combine closed-loop model-based control with distilled contact planning,
our agents display robustness to \textit{sim2real} transfer. Additionally, we note higher puck speeds in the real setting for all agents as a result of differences in real and simulated contact dynamics. Example physical shots of all approaches can be found in the supplementary video.

\begin{table}[t]
\caption{Results of the simulated experiments.}
\label{tab:sim_res}
\begin{tabular}{lccc}
\toprule
& Score & Puck Speed $\left[\frac{\text{m}}{\text{s}}\right]$ (mean $\pm$ std.) & Num. Banks (mean) \\ \midrule
\multicolumn{1}{l|}{CB} & 0.51      & 0.52 $\pm$ 0.24      & 0.00 \\
\multicolumn{1}{l|}{Atacom} &  0.90  &  0.55 $\pm$ 0.05      & 0.00\\
\multicolumn{1}{l|}{Ours \#1} & \bf{0.93} & 1.00 $\pm$ 0.20      & 0.00 \\
\multicolumn{1}{l|}{Ours \#2} & 0.80      & 1.44 $\pm$ 0.63      & 0.53 \\
\multicolumn{1}{l|}{Ours \#3} & 0.61      & \bf{1.97} $\pm$ 0.49 & \bf{1.13} \\ \bottomrule
\end{tabular}
\end{table}

\subsection{Ablation Studies}

\rssparagraph{Ablation study on reduced action space}
We conduct an ablation study on the impact of learning reduced actions, i.e. learning shooting angles (1 DoF) as described in Sec.~\ref{subsec:action_parametrzation}, compared to learning full actions, i.e. the mallet state at impact (4 DoF). Table \ref{tab:sim_res} reports numerical results on the shooting performance using the same number of demonstrations for both modes. It shows that our online inference algorithm with warm-starting also works with higher-dimensional action spaces. However, a higher-dimensional action space requires more data to achieve similar performance.

\begin{figure}[t!]
  \centering
  \includegraphics[width=\linewidth]{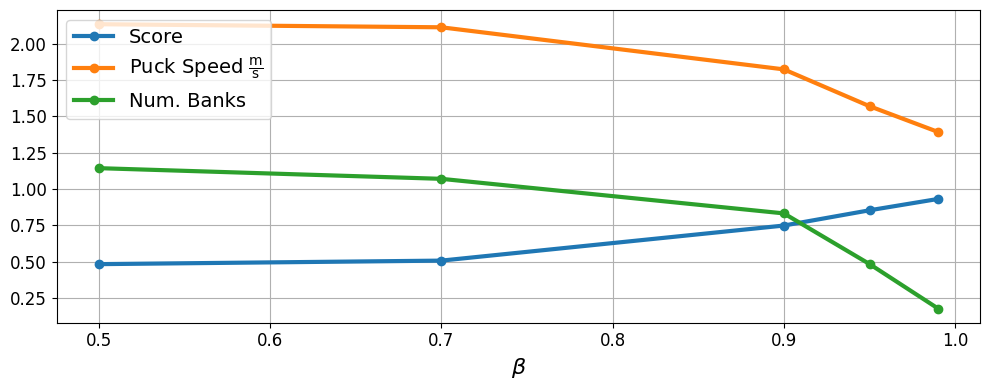}
  \caption{Impact of the chance constraint threshold $\beta$ on shooting performance.\label{fig:constraint}}
\end{figure}

\rssparagraph{Ablation study on chance constraint}
We furthermore conduct a study on the impact of the threshold $\beta$, which constrains the likelihood of hitting the goal as described in Sec.~\ref{subsec:soc}. Fig.~\ref{fig:constraint} illustrates how the shooting behavior changes for various $\beta$ when optimizing for puck speed ($\lambda_1{=}0$, $\lambda_2{=}1$). The score increases with $\beta$, while the main sources of uncertainty, puck speed and number of banks, are reduced with an increasing $\beta$.

\section{Limitations}
Several contributing factors enable the highly performative behavior displayed by our robot air hockey agent. Firstly, the dimension of the task space is low, as the puck is constrained to a plane. While this makes our results relevant for many pushing tasks, most real-world manipulation tasks involve higher-dimensional task spaces.
Along the same line, we were able to leverage existing physical models as a strong prior on contact dynamics. Such models are not available for tasks that involve more complex contact dynamics.
While our approach enables learning on real-world data, we exploited a physics engine to collect data for learning a compact contact model.

\section{Conclusion}
This paper investigated the combination of learning-based contact planning with model-predictive robot control to produce agile behavior in Robot Air Hockey. We show that distilling an optimal contact planning policy through behavior cloning effectively reduces the horizon required for lower-level trajectory optimization, enabling real-time operation.
We show that the proposed approach is capable of accommodating different desired behaviors and sampling-based optimization schemes.
Our results show that the proposed framework outperforms a purely control-based approach and a purely learning-based approach in simulated and real-world games of robot air hockey.
Future work will seek to further leverage the sample efficiency of structured dynamics models to capture the underlying contact dynamics of physical systems from real data. Additionally, integrating physically-informed priors into the implicit model is an interesting direction for increasing the data efficiency of our approach.
Lastly, we are interested in investigating the applicability of our approach to higher-dimensional task spaces with more complex contact interactions.

\begin{table}[t]
\caption{Results of the real-world experiments.}
\label{tab:phys_res}
\begin{tabular}{lccc}
\toprule
& Score & Puck Speed $\left[\frac{\text{m}}{\text{s}}\right]$ (mean $\pm$ std.) & Num. Banks (mean) \\ \midrule
\multicolumn{1}{l|}{CB} & 0.49      & 1.09 $\pm$ 0.24      & 0.00 \\
\multicolumn{1}{l|}{Atacom} & 0.13      & 0.66 $\pm$ 0.15      & 0.31 \\
\multicolumn{1}{l|}{Ours \#1} & \bf{0.78} & 1.72 $\pm$ 0.20      & 0.00 \\
\multicolumn{1}{l|}{Ours \#2} & 0.60      & 2.02 $\pm$ 0.35      & 0.37 \\
\multicolumn{1}{l|}{Ours \#3} & 0.31      & \bf{2.37} $\pm$ 0.50 & \bf{0.90} \\ \bottomrule
\end{tabular}
\end{table}

\section*{Acknowledgments}
This work was supported by the Swiss National Science Foundation (SNSF) through the CODIMAN project, by the State Secretariat for Education, Research and Innovation in Switzerland for participation in the European Commission’s Horizon Europe Program through the INTELLIMAN project (https://intelliman-project.eu/, HORIZON-CL4-Digital-Emerging Grant 101070136) and the SESTOSENSO project (http://sestosenso.eu/, HORIZON-CL4-Digital-Emerging Grant 101070310), by the China Scholarship Council (Grant 201908080039) and by the German Federal Ministry of Education and Research (BMBF) through the KIARA project (Grant 13N16274).

\begin{table}[t]
\caption{Reduced Action Space (R. Act.) v. Full Action Space (F. Act.).}
\label{tab:sim_res}
\begin{tabular}{lccc}
\toprule
& Score & Puck Speed $\left[\frac{\text{m}}{\text{s}}\right]$ (mean $\pm$ std.) & Num. Banks (mean) \\ \midrule
\multicolumn{1}{l|}{R. Act.} & \bf{0.80} & \bf{1.44} $\pm$ 0.63      & \bf{0.53} \\
\multicolumn{1}{l|}{F. Act.} & 0.51      & 0.96 $\pm$ 0.13 & 0.00 \\ \bottomrule
\end{tabular}
\end{table}

\bibliographystyle{plainnat}
\bibliography{main}

\appendices

\end{document}